# Implementation and analysis of Ryze Tello drone vision-based positioning using AprilTags

Kacper Hulek
Warsaw University of Technology
Św. A. Boboli 8, Warsaw, Poland

Mariusz Pawlicki
Warsaw University of Technology
Św. A. Boboli 8, Warsaw, Poland

Adrian Ostrowski
Warsaw University of Technology
Św. A. Boboli 8, Warsaw, Poland

Jakub Możaryn
Warsaw University of Technology
Św. A. Boboli 8, Warsaw, Poland

*Abstract*— **The paper describes of the Ryze Tello drone to move autonomously using a basic vision system. The drone's position is determined by identifying AprilTags' position relative to the drone's built-in camera. The accuracy of the drone's position readings and distance calculations was tested under controlled conditions, and errors were analysed. The study showed a decrease in absolute error with decreasing drone distance from the marker, a little change in the relative error for large distances, and a sharp decrease in the relative error for small distances. The method is satisfactory for determining the drone's position relative to a marker.**

*Keywords— Drone Tello, Computer Vision, MATLAB, Positioning System*

## I. Introduction

Drones are becoming increasingly popular in various fields, such as photography, surveying, transportation, and military systems. The precise positioning of drones is essential for their safe and accurate movement. The GPS is one of the primary positioning systems used for drone flight and is widely used in the field. However, GPS accuracy is significantly reduced in closed spaces.

Our project aimed to develop a solution that only requires a simple vision system to achieve accurate positioning in closed spaces. Our solution eliminates the need for additional hardware and involves providing good lighting and using standardized AprilTags to mark the places where the drone needs to move. We used the "tag36h11" family tag to ensure real-time performance while minimizing false-positive detections.

In the following sections, we discussed the technologies we used and the preparation of the drone workspace required for the program to function correctly. We also presented the operating procedure of our application, including a discussion of the graphical user interface available to the user. Finally, we included an error analysis of the drone's accuracy in determining its position relative to the marker.

## II. Implementation details

### A. Required hardware

The Tello drone is a small quadcopter with a vision positioning system and an onboard camera. It can hover in place using a vision positioning system and an advanced flight controller. It has advanced features such as Bounce mode, 8D Flip and EZ Shots. The drone takes 5 MPx photos and streams live video in 720 p resolution. Maximum flight length reaches up to 100 meters.

The Ryze Tello drone has a dedicated Support Package for Ryze Tello Drone [2] library in the MATLAB environment allowing for easy control, image capture and video upload (Table 1, Fig. 1). In addition, the library also has functions to collect data about its location, speed, and battery status. The capabilities of the library, and the functions we used, are described in more detail in the next section.

TABLE 1. RYZE TELLO DRONE PARAMETERS

| Flight time | Maximum flight speed | Battery capacity | Weight | Camera resolution |
|---|---|---|---|---|
| 13 min | 8 m/s | 1100 mAh | 80 g | 960 / 720 px |

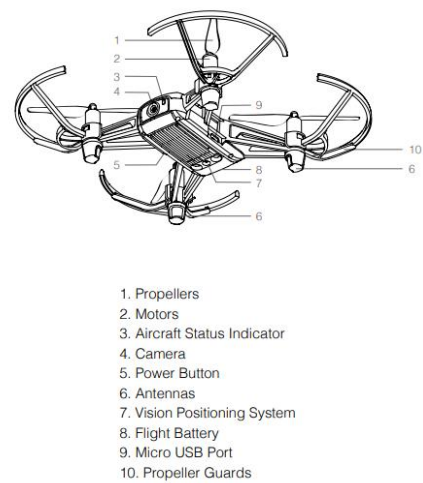

Fig. 1. Construction of the Ryze Tello drone. [1]

### B. Required software

The first part of the project involved developing a functionality to detect AprilTags and determine their location. To accomplish this, we used the ReadAprilTag library [3], part of the Computer Vision Toolbox [4] in the MATLAB environment. Initially, we needed to establish the camera parameters to determine the actual displacement of the tags relative to the captured image.

To obtain high-quality images, we chose to use the `snapshot()` function from the Support Package for Ryze Tello Drone library, which enables the capture of a single image from the drone's camera. Although streaming video using the preview() function is possible, the video quality in streaming mode is noticeably worse. It might need to be improved to determine the location of markers accurately. A Wi-Fi module is required to connect to the drone and run the program remotely.

III. RYZE TELLO DRONE POSITIONING PROCEDURE

A. *Determining the camera parameters of the Ryze Tello drone*

To read the position of AprilTag tags correctly, it is necessary to define the actual size of the tag and then specify the camera parameters. In our tests, tags with a side size of 184 mm were used, but the created program allows to work with tags of any size. In this case, the user must specify a custom size in the application window.

To enable the tag detection algorithm to work, the camera parameters required are the resolution of the camera image (ImageSize[x, y]), specified in two-axis parameters in pixels, the center point of the sensor (PrincipalPoint[x, y]), also given in two-axis parameters in pixels, and the axial focal length (FocalLength[x, y]), also specified in two-axis parameters in pixels.

For the Ryze Tello drone, the ImageSize value documented equals 960 by 720 pixels. Assuming ideal sensor performance, we can assume that the PrincipalPoint parameter specifies the point exactly in the center of the image. Therefore, we can take half of the ImageSize parameter's value as its value.

Regarding the FocalLength parameter, the readAprilTag [5] library documentation states that it is the product of the value of the primary focal length of the camera and the quotient of the resolution of the maximum working area of the sensor and the physical size of this area in a given axis (Fig. 2). We can determine this value using the following formula

$$f_{x,y} = f \cdot \frac{r_{x,y}}{s_{x,y}} \quad (1)$$

Where $f_{x,y}$ is the desired axial focal length, $f$ is the normal focal length value of 4mm for this camera, $r_{x,y}$ is the resolution along a given axis in pixels, and $s_{x,y}$ is the sensor dimension along a given axis in mm.

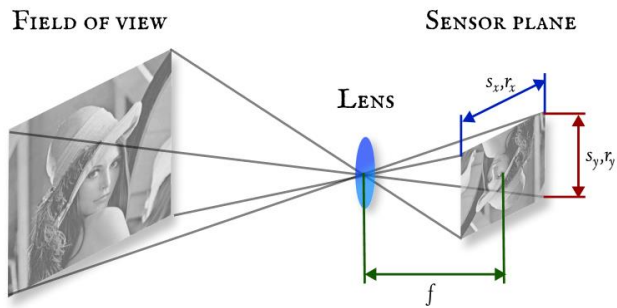

Fig. 2. Designation of the quantities needed to calculate the focal values.

Images from the drone's camera are captured using the snapshot() function, and then to get rid of image distortion, including mainly the fisheye effect causing angular warping at the edges of the image, the undistortimage() function, a part of the Computer Vision Toolbox library, is called (Fig. 3).

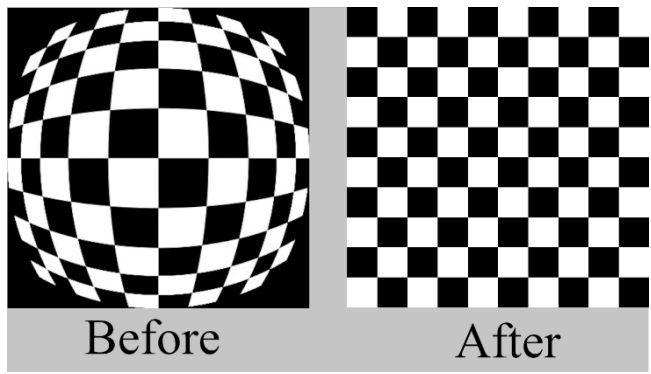

Fig. 3. Example comparisons of the photo before distortion correction (left), and after (right).

B. *Preparation of the flight environment*

It is crucial to ensure proper flight conditions to fly a Ryze Tello drone based on positioning relative to AprilTag tags. The drone comes equipped with a position stabilization system with a vision system consisting of cameras built into the device aimed at the ground. This system is designed to keep the drone in a fixed position relative to the ground (Fig. 4).

To ensure that the system works properly, it is essential to ensure that the ground pattern is not uniform and is as contrasting as possible. A soft substrate can also minimize the airflow created by the drone's rotors, which can cause unwanted movement of the device. Additionally, the flight environment should be well-illuminated to enable the correct identification of the markers. These measures are crucial to ensure the proper functioning of the drone's vision system and accurate positioning relative to the AprilTag markers.

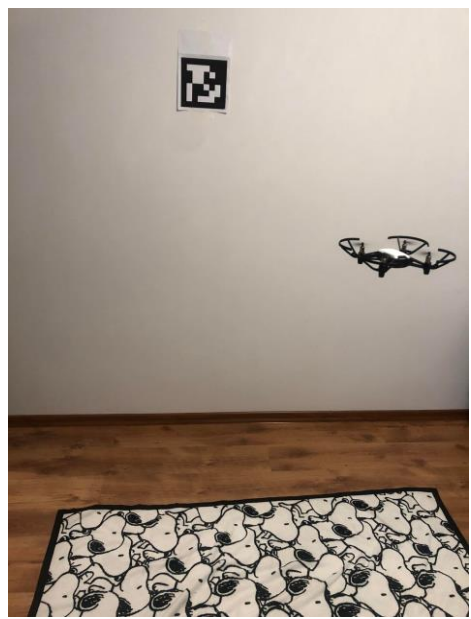

Fig. 4. Ryze Tello drone in correctly prepared flight environment.

## C. The process of identifying the location of AprilTags

To implement AprilTag detection, the `readAprilTag()` function from the readAprilTag library [3] is used. This function returns a pose object containing the detected tags' position in the dimensionless units used in the library. An empty pose object is returned if no marker is detected in the image.

The `Translation()` method is used on this pose object to obtain the actual offset of the markers relative to the camera in meters. This method returns vectors that represent the position of the markers relative to the point [0, 0, 0] that defines the camera. These vectors contain [x, y, z] coordinates corresponding to rightward, downward, and forward displacement. Using this method, the precise location of the markers relative to the camera can be determined and used for accurate drone positioning.

## D. The process for autonomous movement of the Ryze Tello drone

To enable control and measurement of the Ryze Tello drone, we utilized the Support Package for the Ryze Tello Drone library. This library provided the means to control the drone's movement and obtain real-time altitude and battery status parameters through a graphical interface. To establish a connection between the drone and the program, we used the `ryze()` function after connecting the drone to the computer via wifi. We also created a camera object using the `camera()` function to facilitate image capture.

We employed the `move()` and `turn()` functions for drone movement. In the `move()` function, the [x, y, z] coordinates correspond to forward, right, and downward movement. However, this coordinate system differs from the one returned by the `readAprilTag()` function (Fig. 5). To ensure proper drone control, we transformed the `readAprilTag()` function vector as follows

$$[x, y, z] = [z', x', y'] \quad (2)$$

Where $x,y,z$ are the values sent to the drone's move() function, and $x',y',z'$ are the values received from the readAprilTag() function.

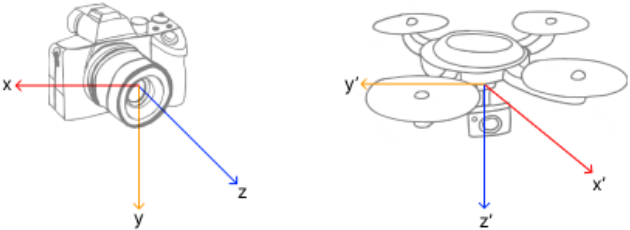

Fig. 5. Comparison of coordinate systems from the perspective of the drone's move() function (left) and from the perspective of the readAprilTag() function after returning the tag move vector (right).

The `turn()` function performs rotation by a given angular value expressed in radians clockwise. When writing the algorithm, attention was paid to ensuring that the value of the calculated needed rotation angle followed the convention described for this function. This angle was obtained by dropping the normal vectors of the camera and the marker into the horizontal plane by replacing the "z" coordinates with "0" values and then calculating the angle between the two vectors using the following formula

$$\theta = cos^{-1}\left(\frac{a \cdot b}{|a||b|}\right) \quad (3)$$

Where $\theta$ is the angle between the vectors in radians, $a$ and $b$ are the normal vectors of the camera and a marker

Knowing that after making the drone rotate by this angle in the opposite direction to the one obtained, it would be perpendicular to the marker, it was now still necessary to transform the movement vector to the marker by the rotation angle made. To do this, the coordinate system of the movement vector was converted to the polar system, and then the rotation angle was subtracted. The result of this operation was then converted back into a Cartesian vector, which became the new movement vector.

In the flight process, a `takeoff()` function was also used to realize the ascent of the drone to an initial height of about 80cm and a `land()` function to allow the drone to land safely from whatever height it is at the given moment. The sequence of operations performed during the simple algorithm for positioning oneself relative to the marker was, therefore as follows:

1) Establishing a connection with the drone and creating a camera object.

2) Specifying the camera parameters and the size of the marker.

3) Launch the drone using the takeoff() function.

4) Capturing the photo using the snapshot() function.

5) Removing the distortion of the photo using the `undistortimage()` function.

6) Performing a tag position calculation using the `readAprilTag()` function.

7) Rotating the drone using the `turn()` function by the opposite value of the camera and tag normal angles in order to position the drone perpendicular to the tag

8) Converting the measured displacement vector by the calculated rotation angle.

9) Executing a move using the `move()` function by the transformed move vector with the forward movement distance value subtracted by the user-specified value of the final distance of the drone in front of the marker.

10) Performing a landing using the `land()` function.

In the case of this project, the drone performed actions 4 through 9 in a loop, returning to the starting position in the horizontal plane after performing a certain action. After failing to detect a marker or returning to the initial position after performing a certain action, the drone would perform a 45-degree turn to find the next marker. After performing a full turn, it would fly up by the height specified by the user in the program and start the operations again until it exceeded the maximum height. In our case, the values were 0.5 and 2 meters, respectively.

The other functions used in the project referred to the abovementioned analysis of the drone's positioning with the `readHeight()` function reading the height from takeoff, `readOrientation()` reading the drone's position relative to the takeoff in the form of Euler angles, `readSpeed()` reading the drone's speed, and the `BatteryLevel()` method for reading the drone battery status value.

## IV. USE OF GRAPHICAL INTERFACE TO CONTROL THE FLIGHT PROCEDURE

The graphical user interface was created using the App Designer tool provided by MATLAB (Fig. 6).

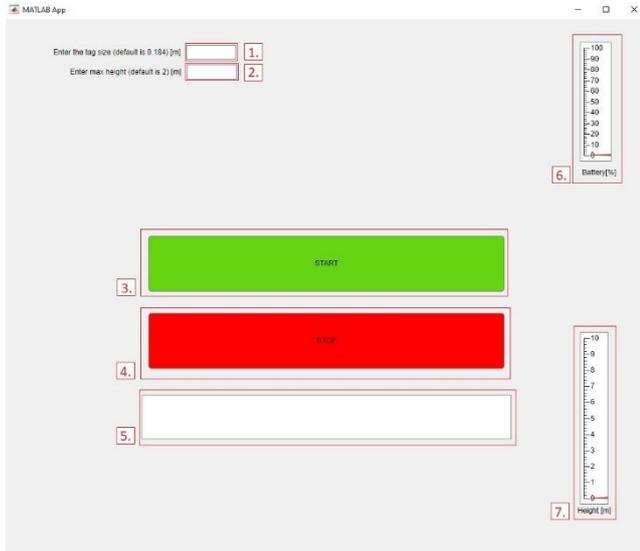

Fig. 6. Graphical interface with the following elements: (1.) a box for entering the size of the side of the marker, (2.) a box for entering the maximum flight altitude, (3.) a button to start the execution of the marker search program and display the camera preview and the flight trajectory, (4.) a button to stop the program, forcing the drone to land, (5.) a box in which messages are displayed when certain actions are not performed, (6.) a battery status indicator expressed in percent, (7.) a drone altitude indicator expressed in meters.

With the interface, one can establish designated parameters, initiate or cease the program, and retrieve crucial flight analysis information (Fig. 6).

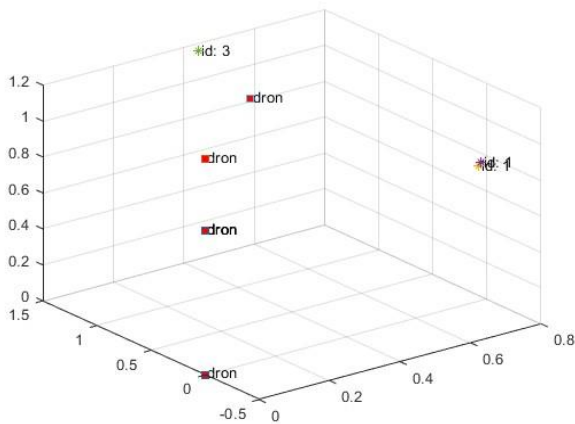

Fig. 6. An example of a drone's flight trajectory and the detected markers. Distances are given in meters.

## V. ERROR ANALYSIS OF THE VISION-BASED POSITIONING METHOD

### A. Method of determining the error

A series of tests were conducted indoors with consistent lighting to verify the accuracy of the drone's position readings and distance calculations. External factors that could affect the algorithm's accuracy were excluded. The measurements were performed using a Bosch DLE 40 laser rangefinder with a precision of +/- 0.0015 meters.

The test involved flying the drone in a straight line towards a marker at a fixed height, taking pictures using the drone's camera, and then measuring the distance to the marker's centre from the camera's location using the rangefinder. We then used a previously written function in MATLAB to read the marker's position in subsequent photos, which returned the position as a vector [x, y, z], where x represents the left/right shift, y represents the up/down shift, and z represents the forward shift (into the photo).

The distance to the marker was defined as the length of the resultant vector [x, y, z]. Using the value measured with the rangefinder as a reference reading, we calculated the absolute and relative errors of reading the marker's offset relative to the camera using the method employed in this project.

### B. Analysis of positioning errors

After conducting a series of measurements, moving the drone away from the marker position, we obtained the following characteristics of the absolute error in the vector line drawn from the camera position to the centre of the marker.

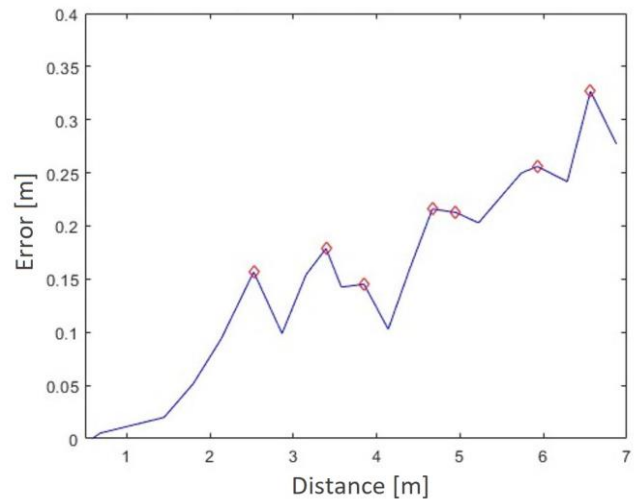

Fig. 7. Changes of the absolute error during the drone flight.

The graph above shows a decrease in absolute error with decreasing drone distance from the marker (Fig. 7). Ideally, the characteristic would be linear. Still, many points appear that reject this hypothesis. Nevertheless, the error trend is downward, which allows the drone's position to improve as it approaches the marker.

Analysing the graph of relative error (Fig 8.), there is little change for large distances, that is, in the range of 7 to 3 meters of distance from the marker. The error fluctuates between 5.5 and 3 per cent. For small distances, i.e., less than 2 meters, a sharp decrease occurs, below 3 per cent error, and at distances of less than 1 meter, the error comes as low as 0.5 per cent. Over its entire range, the relative error does not exceed 7%, which is a very satisfactory result.

To analyse the origin of the errors in more depth, we have marked the points for which increases are visible on the graphs concerning the previous measurement (marked with red diamonds). Our considerations are conducted assuming the drone is getting closer to the markers. We can observe that the increases appear mainly for measurements taken at a distance of more than 3 meters. They may be due to the measurement inaccuracy of the laser meter operator, the drone's camera's low resolution, the drone's camera, or the angular offset of the drone relative to the marker. For the relative error, changes at large distances do not matter, while the closer the marker is to the camera, the greater the impact of the disturbance of the absolute error; that is, despite small increases in the absolute error, the relative error visibly increases.

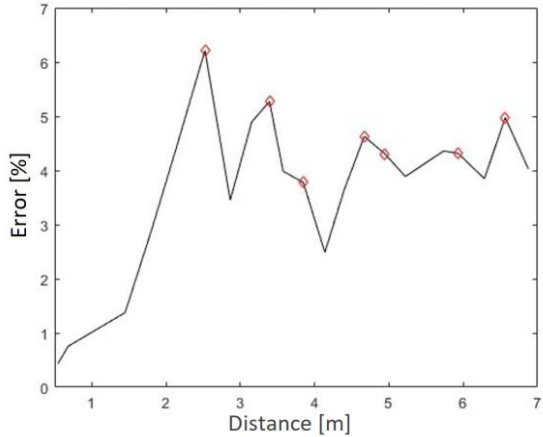

Fig. 8. Changes of relative error during the drone flight.

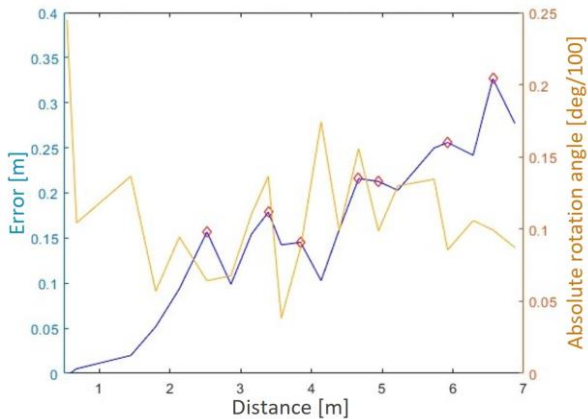

Fig 9. The effect of the angular offset of the drone.

The final element of this analysis is to examine the effect of the angular offset of the drone, relative to the marker, on distance reading errors (Fig. 9). To do this, a line representing the change in angular displacement was plotted on the absolute error graph. The angular values are placed on the right axis at a scale of 1 : 100. As can be seen from the graph, the error values cannot be related to the angular displacement values. The data appear to be independent.

## VI. SUMMARY

The goal of positioning the drone relative to the markers has been achieved. Based on the analysis of the measurement error of the method used in the project, it can be seen that the angular displacement of the marker does not affect the accuracy of determining its position relative to the camera. In addition, the data presented show that when determining the position of the marker at distances of less than 3 meters, the relative error decreases significantly and reaches values of about 0.5%, which allows for very accurate motion correction in the final stage of positioning. On the other hand, at distances of more than 3 meters, the relative error reading remains constant at between 3 and 7 percent. These types of readings allow for satisfactory positioning accuracy of the drone in this type of application.

The examined method shows promising results in exemplary testing although further development is necessary to achieve consistent outcomes. In the case of continuous marker detection through video transmission, it would be possible to obtain more accurate positioning results due to continuous position correction. However, this solution requires a more accurate video capture quality.